%% file: main.tex
\pdfoutput=1
\documentclass[11pt]{article}

\usepackage[]{emnlp2021}

\usepackage{times}
\usepackage{latexsym}
\usepackage[T1]{fontenc}
\usepackage[utf8]{inputenc}
\usepackage{microtype}

\usepackage{array}
\usepackage{multirow}

\usepackage[T1]{fontenc}

\usepackage[utf8]{inputenc}
\usepackage{textcomp}

\usepackage{graphicx}

\usepackage{subcaption}
\usepackage{url}
\usepackage{amsmath, amsthm, amssymb}

\usepackage{paralist}
\usepackage{xspace}
\usepackage{booktabs} 
\usepackage{wasysym} 

\usepackage{tikz}
\usetikzlibrary{calc}
\usepackage{adjustbox}
\usepackage{floatrow}
\usepackage{float}
\usepackage{arydshln}
\usepackage{centernot}
\usepackage{bbm}

\usepackage{circledsteps}
\newcommand\myCircled[2][]{\ifmmode
\Circled[fill color=black,inner color=white,#1]{\textbf{\mathsf{#2}}}
\else
\Circled[fill color=black,inner color=white,#1]{\textbf{\sffamily#2}}
\fi
}

\newcommand{\alphaOne}{$\alpha_1$\xspace}

\newcommand{\trainSet}{$Train$\xspace}

\newcolumntype{P}[1]{>{\centering\arraybackslash}p{#1}}

\usepackage{pdfpages}

\newcommand{\platformName}{{\sc TabPert}\xspace}

\newcommand{\skippedDetails}[1]{}

\usepackage{xcolor}

\definecolor{orange}{rgb}{1,0.5,0}
\definecolor{mdgreen}{rgb}{0.05,0.6,0.05}
\definecolor{mdblue}{rgb}{0,0,0.7}
\definecolor{dkblue}{rgb}{0,0,0.5}
\definecolor{dkgray}{rgb}{0.3,0.3,0.3}
\definecolor{slate}{rgb}{0.25,0.25,0.4}
\definecolor{gray}{rgb}{0.5,0.5,0.5}
\definecolor{ltgray}{rgb}{0.7,0.7,0.7}
\definecolor{purple}{rgb}{0.7,0,1.0}
\definecolor{lavender}{rgb}{0.65,0.55,1.0}

\usepackage{enumitem}

\usepackage{enumitem}

\newcommand{\datasetName}{{\sc InfoTabS}\xspace}

\title{\platformName : An Effective Platform for Tabular Perturbation}

\author{Nupur Jain \\
   IIT Kanpur\\
  \texttt{nupurj@iitk.ac.in} \\\And
  Vivek Gupta \\
  University of Utah \\
  \texttt{vgupta@cs.utah.edu} \\ \AND
  Anshul Rai \\
   IIT Kanpur \\
  \texttt{anshulra@iitk.ac.in} \\  \And 
    Gaurav Kumar \\
  IIT Kanpur \\
  \texttt{gauravkg@iitk.ac.in} \\
  }
  
\begin{document}
\maketitle

\begin{abstract}
\input{Abstract}
\end{abstract}

\input{introduction}

\input{motivation}

\input{tabpert_features}

\input{casestudy_infotabs}

\input{utility}

\input{future_work}

\input{conclusions}

\bibliography{anthology,custom}
\bibliographystyle{acl_natbib}

\clearpage
\newpage

\appendix
\label{sec:appendix}
\input{table_perturb_stats}
\input{comparision}

\input{examples}

\end{document}

%% file: abstract.tex
To truly grasp reasoning ability, a Natural Language Inference model should be evaluated on counterfactual data. \platformName facilitates this by assisting in the generation of such counterfactual data for assessing model tabular reasoning issues. \platformName allows a user to update a table, change its associated hypotheses, change their labels, and highlight rows that are important for hypothesis classification. \platformName also captures information about the techniques used to automatically produce the table, as well as the strategies employed to generate the challenging hypotheses. These counterfactual tables and hypotheses, as well as the metadata, can then be used to explore an existing model's shortcomings methodically and quantitatively.

%% file: introduction.tex
\section{Introduction}
\label{sec:introduction}

Given the factual evidence, a crucial part of NLP model reasoning capacity is if it can evaluate whether a given hypothesis is an entailment (true), contradiction (false), or neutral (can't be determined). Current transformers-based models have been shown to outperform humans on these tasks when the evidence is presented as simple unstructured text \cite{wang2018glue,wang2019superglue}; however, when tested with semi-structured evidence \cite{gupta-etal-2020-infotabs,chen2019tabfact}, such as tables, as shown in Figure \ref{fig:example}, the very same models struggled to match human accuracy \cite{neeraja-etal-2021-incorporating,wang2021semeval,aly2021feverous}.

Furthermore, there can be several reasons for a model's correct predictions on a particular example. For example, \citet{poliak-etal-2018-hypothesis,gururangan-etal-2018-annotation} show that multiple NLI datasets, such as SNLI and MNLI dataset \cite{snli:emnlp2015,williams-etal-2018-broad} exhibit hypothesis-bias, i.e., the hypothesis-only model performs significantly better than the majority label baseline. In the context of tables, \citet{gupta-etal-2020-infotabs} shows that the right prediction doesn't always imply reasoning: there can be dataset biases in semi-structured datasets too, such as hypothesis or premise artifacts (spurious patterns) which can wrongly support a particular label. 

Furthermore, a model can also ignore the ground evidence and use its pre-trained knowledge for making predictions. These models when deployed in the real world on out-of-domain (different category) or counterfactual (stories tables) examples fail embarrassingly. One way to avoid this inflated performance projection is to test models on several challenging sets before actual deployment. For example, \citet{gupta-etal-2020-infotabs} evaluate the RoBERTa$_{(Large)}$ \cite{liu2019roberta} models on two additional adversarial sets (hypothesis perturbed and out-of-domain) and observe a significant performance drop. However, manually creating such challenge sets can be tricky, both in terms of the annotation cost involved and the actual annotation process, especially with tabular data of semi-structured nature. 

\begin{figure}
  \centering
  {
  \footnotesize
  \begin{center}
    \begin{tabular}{>{\raggedright}p{0.25\linewidth}p{0.6\linewidth}}
      \toprule
      \multicolumn{2}{c}{\bf New York Stock Exchange}                                                    \\
      \midrule
      {\bf Type} & Stock exchange \\
                                              
      {\bf Location}                & New York City, New York, U.S.                                                  \\ 
      {\bf Founded}           & May 17, 1792; 226 years ago          \\  
      {\bf Currency}                  & United States dollar                             \\ 
      {\bf No. of listings}      & 2,400                                            \\  
     {\bf Volume}             & US\$20.161 trillion (2011)                                                 \\ 
      \bottomrule
    \end{tabular}
  \end{center}
}
  {\footnotesize
    \begin{enumerate}[nosep]
    \item[H1:] NYSE has fewer than 3,000 stocks listed.
    \item[H2:] Over 2,500 stocks are listed in the NYSE.
    \item[H3:] S\&P 500 stock trading volume is over \$10 trillion.
    \end{enumerate}
    }
    \caption{A tabular premise example. The hypothesis H1 is entailed by it, H2 is a contradiction and H3 is neutral i.e. neither entailed nor contradictory.}
  \label{fig:example}
\end{figure}

Recently, \citet{ribeiro2020beyond} showed that one can deploy simple tricks to semi-automate this process and develop several adversarial counterfactual contrast sets by altering existing data to perform behavioral testing of a model. However, such tricks currently only work for unstructured text and cannot be directly adopted for semi-structured text such as tables. To fill this gap, in this work, we present \platformName. \platformName which is an annotation platform especially designed to work on semi-structured tabular data. \platformName support semi-automatic creation of tabular counterfactual data. Through \platformName annotators can modify tables in several ways such as \begin{inparaenum}[(a)] 
\item \textit{deleting information}: deleting an attribute-value pair or an existing row completely, 
\item \textit{inserting information}: inserting an attribute-value pair for an existing row or creating a fresh row, 
\item \textit{modifying information}: editing the attribute or values cells of an existing row, and
\item \textit{modifying hypothesis or label}: modifying an existing hypothesis and its inference label.
\end{inparaenum}

\platformName also automatically logs the modification operation for each attribute-value of the table with respect to the original table. Furthermore, users can manually log information about the relevant rows and the strategy used for perturbing a table-hypothesis pair, in addition to the gold label, through \platformName. Such metadata is very important in assessing annotated data toughness and can be later utilised to systematically study failure modes of a model.

The contributions of our work can be summarised as below:
\begin{enumerate}
     \item \platformName can delete, modify, and insert information in semi-structured tabular data for creating counterfactual examples
    \item \platformName auto-logs table perturbation metadata, and support manual hypothesis  modification and inference labels selection.
    \item \platformName assists users in logging metadata including hypothesis-related rows and the perturbation strategy used, which is useful for data and model quantitative analysis
    \item We present a case study for \platformName via the generation and  evaluation of a counterfactual \datasetName dataset and models respectively. 
\end{enumerate}

The \platformName source code, the annotated counterfactual \datasetName dataset, along with the RoBERTa$_{Large}$ model, the annotation instructions and examples set, and all other associated scripts, are available at \url{https://github.com/utahnlp/tabpert}. The instruction video describing \platformName usage is accessible at \url{https://www.youtube.com/watch?v=sbCH_zD53Kg}.

%% file: motivation.tex
\section{Tables are Challenging}
\label{sec:motivation}

One might argue that creating a counterfactual dataset for tables is not a challenging task, and that table modification can be fully automated by merely \emph{`shuffling' or `inserting'} attribute values of one table row into another table row (with the same attribute) as long as they are from similar categories, e.g. shuffle `producer' of one film with `producer' of another film). One can extend this further by shuffling rows with different attributes in the same as well as different tables (same category) as long as the name-entity type for values are similar, e.g. shuffle `producer' with the `director' of the same or a different film with each other.

However, this approach does not automate the modification of corresponding hypotheses and their inference labels. Furthermore, such automatic shuffling encourages the violation of certain natural common-sense logical constraints, such as a person's `Birth Date' must be before their `Died Date', a person's `Marriage Date' should be after their `Birth Date' and before their `Died Date', an album's `Released Date' should be after its `Recording date' and so on. Without the enforcement of these constraints, the table will be self-contradictory. While some of these constraints can be automatically satisfied and hence not violated, a majority still slip through because of their sheer variety and variation. \footnote{In real data, these constraints are naturally satisfied.}. Furthermore, enforcing these constraints automatically during perturbation is a challenging job due to its domain-specific nature. However, such automatic perturbations can be a good initialization for our \platformName tables, which can then be manually inspected and modified by human annotators for self-consistency i.e. no natural common sense violation.

%% file: tabpert_features.tex
\section{\platformName Functions, Aspects, and Usability }
\label{sec:func_aspect_usage}

\platformName is currently supported on common web browsers such as Google Chrome and can be installed to run locally \footnote{Details: \url{https://github.com/utahnlp/tabpert}}. There are three main steps required for successful annotation, as described below.

\subsection{Automatic Initialization}

First, we initialise \platformName with original tables and automatic counterfactual tables generated via automatic `shuffling' of table rows or attribute values. Automatic initialization is beneficial as manual table creation is both time-consuming and highly error-plausible. This automatic shuffling operation is stored as a metadata of each attribute-value in the first 4 bits of a 7-bit string, and can be used later to analyse which shuffling was more effective. Table \ref{tab:fourbits} shows the meaning of each of these bits \footnote{For same table i.e. $3^{rd}$ bit $0$, the $1^{st}$ and $2^{nd}$ bits are zero}. Counterfactual hypothesis (and label) initialization is done by copying the information from the original table exactly.

\begin{table}
\small
\centering
\begin{tabular}{c|c|c|c}
\toprule
\bf Bit &\bf Location &\bf Same &\bf Different\\
\midrule
1 & Dataset & 0 & 1 \\
2 & Category & 0  & 1 \\
3 & Table & 0 & 1 \\
4 & Key & 0 & 1 \\
\bottomrule
\end{tabular}
\caption{\small First Four Bits of Table Value Metadata. The $1^{st}$ bit represent the perturb value table location i.e. whether it is coming from the Train set (1) or the Test set (0), the $2^{nd}$ bit indicates the different (1) or same (0) category, the $3^{rd}$ bit represents if the value is from the same (0) or from different (1) table (for same table, $1^{st}$ and $2^{nd}$ bit is always zero), and the $4^{th}$ represents same (0) or different (1) attribute.}
\label{tab:fourbits}
\end{table}

\subsection{Modifying Tables}
The automatically perturbed tables from initialization can now be manually modified to create counterfactual examples\footnote{Also to avoid self-contradiction or inconsistency.}. All the cells (attributes and values) in the three counterfactual tables can be edited\footnote{Original table is fixed to prevent inadvertent edits}. Table rows can be modified via the dragging and dropping of a value cell from \begin{inparaenum}[(a)] \item same counterfactual table (cut-paste effect),  \item from another counterfactual table (cut-paste effect), \item from the original table (copy-paste effect)\end{inparaenum}. To minimise errors during this drag-and-drop operation, a type validation check runs in the background, which prevents a drag and drop between different key categories (for example, it is forbidden to drag a \emph{Person Name} into \emph{Date of Birth}). To achieve such type-check, keys are automatically grouped using category (entity type) information during initialization. \footnote{Keys for which a category information is missing can be dropped without restriction.}

\platformName also supports five additional functions for more challenging edits. The \emph{`Add'} box allows annotators to write text and drag and drop it into a table to create a new value. For deleting a value cell, simply drag and drop the value cell to the \emph{'Delete'} Box. One can also edit the text for an existing attribute or value via \emph{clicking} it. Lastly, one can delete any row with the \emph{`Edit'} option, and also insert a new row and its details using the \emph{'Add Section'} button. These modification details are also recorded automatically in the last 3 bits of the 7-bit metadata. The $5^{th}$ bit represents copy-paste from original, the $6^{th}$ represents a new cell or row addition, and the last $7^{th}$ bit represents a value update operation. Figure \ref{fig:main_platform_table} shows the main parts of the \platformName platform for counterfactual table perturbation.


\begin{figure*}[!htbp]
    \centering
    
    \begin{subfigure}[t]{0.98\textwidth}
        \caption{\textbf{Table Perturbation : } \myCircled{1}Table Title \myCircled{2}Values \myCircled{3}Key (table section name) \myCircled{4} Edit Section Name or Delete Section \myCircled{5}Add New Value \myCircled{6}Delete Value \myCircled{7}Add New Section}
        
        \input{screenshot_table}
        \label{fig:main_platform_table}
    \vspace{-2.0em}
    \end{subfigure}
    \begin{subfigure}[b]{0.48\textwidth}
        \input{screenshot_hypotheses}
        \vspace{-1.5em}
        \caption{ \textbf{Hypotheses Perturbation: } \myCircled{1}Hypotheses Table Labels \myCircled{2}Label of an Original Hypothesis \myCircled{3}Label of a Counterfactual Hypothesis of Table A \myCircled{4} Original Hypothesis \myCircled{5}Counterfactual Hypothesis in Table A \myCircled{6}Option to open Modal for Selecting Hypothesis metadata (Figure \ref{fig:main_platform_relevant_row}) \myCircled{7}Submit Button to Save Work}
        \label{fig:main_platform_hypothesis_perturb}
    \end{subfigure}
    \begin{subfigure}[b]{0.48\textwidth}
        \input{screenshot_strategies}
        \vspace{-1.75em}
        \caption{\textbf{Selection of Relevant Rows and Hypothesis Perturbation Strategies :} \myCircled{1}Hypothesis \myCircled{2}Table Label \myCircled{3}Relevant Row (checkbox ticked) \myCircled{4} Irrelevant Row(checkbox not ticked) \myCircled{5}Strategy Used (checkbox ticked) \myCircled{6}Strategy Not Used (checkbox not ticked)}
        \label{fig:main_platform_relevant_row}
    \end{subfigure}
    \label{fig:main_platform_figure}
\end{figure*}

\subsection{Hypothesis Modification and Metadata}
The text of a hypothesis of a counterfactual table can be edited directly and its corresponding label can also be selected from drop-down menu options. In addition to this, other metadata information is also collected:

\begin{enumerate}
    \item The strategies used by the annotator to modify the hypothesis. The five main strategies can be selected using the multi-value check-box. The `Other' option corresponds to hypothesis changes that don't fall into main strategies.  
    \item All the relevant rows of the table which are necessary for deciding the inference label. 
\end{enumerate}

Figure \ref{fig:main_platform_hypothesis_perturb} shows the main \platformName view for hypothesis modification, with hypothesis and inference label. Metadata is inserted by the annotator via clicking `+' symbol on the left side (below label drop-down) for each hypothesis, as shown in Figure \ref{fig:main_platform_hypothesis_perturb}. This opens a metadata collection window, as shown in \ref{fig:main_platform_relevant_row}). Here too, we use 6 bits: the initial five for each strategy (bits order is the same as the order in which the strategies are mentioned in \platformName as shown Figure \ref{fig:main_platform_relevant_row}), and the last one for the `Other' option. We store the relevant rows `attribute keys' in a list (array) for each hypothesis along with the final hypothesis text modification.

\textbf{\platformName Aspects:} The \platformName web-app’s core tech stack consists of ReactJS\footnote{https://reactjs.org/} and Flask\footnote{https://flask.palletsprojects.com/en/2.0.x/}. Here, Flask is used as the main back-end Python web framework, and javascript library ReactJS is used for the front-end. We used Flask because is easy-to-extend, giving us the ability to easily integrate Python libraries for quick manipulation of \emph{JSON} and \emph{TSV} files. We used ReactJS because of the react-beautiful-dnd library\footnote{https://www.npmjs.com/package/react-beautiful-dnd} essential for simulating the drag-and-drop function.

%% file: screenshot_table.tex
\tikzset{
mystyle/.style={
  circle,
  inner sep=0pt,
  align=center,
  draw=black,
  fill=black,
  text=white,
  minimum size=10pt
  }
}
\begin{adjustbox}{width=\textwidth}

\begin{tikzpicture}[my label/.style n args={2}{label={[font=\small,text=white,xshift=-0.4cm]#1:#2}}]
 
    \node[above right, inner sep=0] (image) at (0,0) {
        \includegraphics[width=\textwidth]{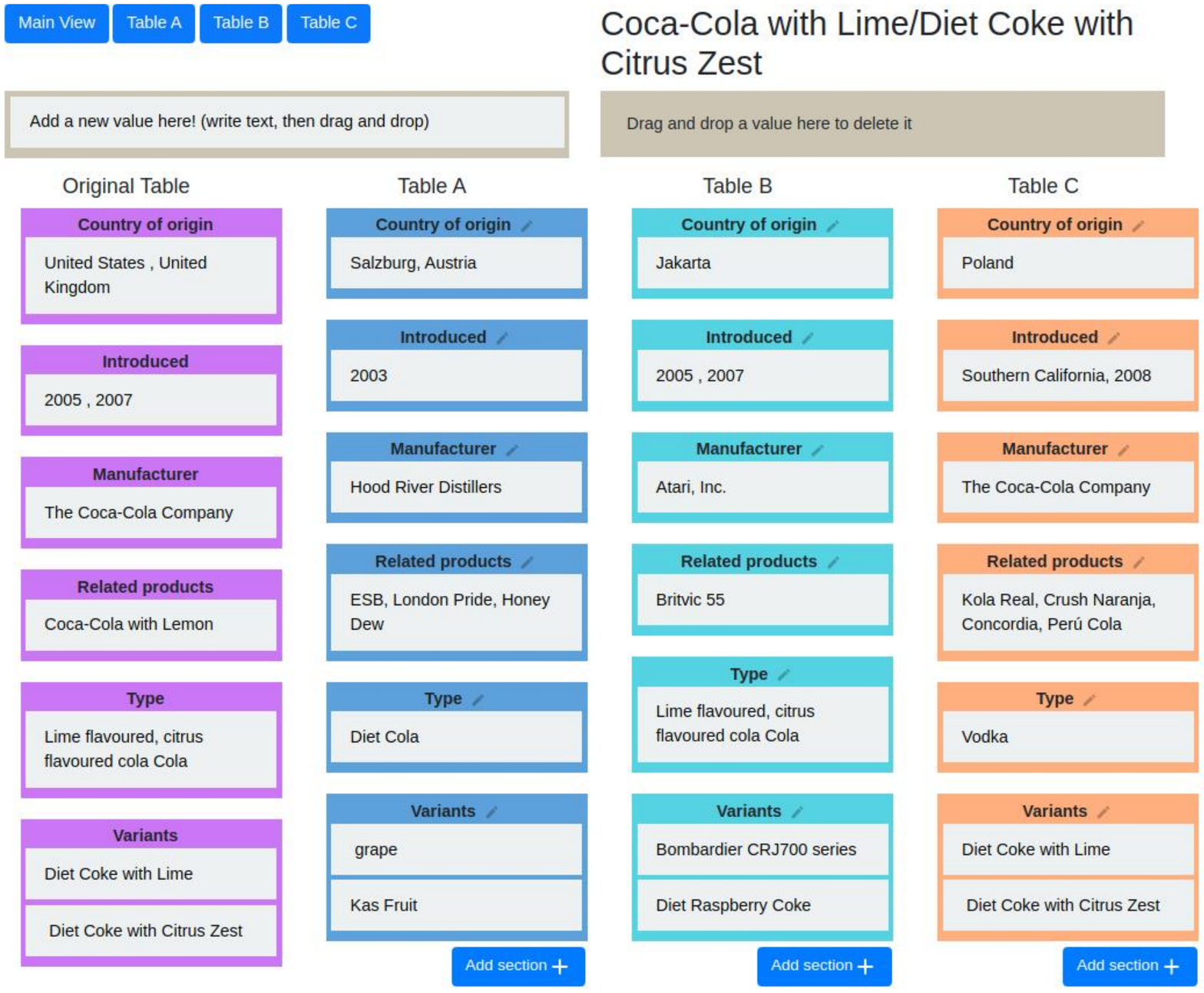}
    };
    \begin{scope}[
    x={($0.1*(image.south east)$)},
    y={($0.1*(image.north west)$)}]
     

    
    \node[mystyle, my label={0}{\textbf{1}}] at (4.8,9.4){};
    \draw[thick,red] (7.2,1) rectangle (9,2) node[below right,mystyle, yshift=-0.35cm,xshift=0.1cm,my label={0}{\textbf{2}}]{};
    \draw[thick,red] (7.5,5.65) rectangle (8.6,5.4) node[below right,mystyle,yshift=0.2cm,xshift=0.1cm,my label={0}{\textbf{3}}]{};
    \draw[thick,red] (4.36,4.3) rectangle (4.55,4.6) node[below right,mystyle,yshift=-0.05cm,xshift=0.1cm,my label={0}{\textbf{4}}]{};
    \node[mystyle, my label={0}{\textbf{5}}] at (4.2,8.6){};
    \node[mystyle, my label={0}{\textbf{6}}] at (8,8.6){};
    \node[mystyle, my label={0}{\textbf{7}}] at (5.8,0.5){};
 
\end{scope}
\end{tikzpicture}
\end{adjustbox}

%% file: screenshot_hypotheses.tex
\tikzset{
mystyle/.style={
  circle,
  inner sep=0pt,
  align=center,
  draw=black,
  fill=black,
  text=white,
  minimum size=7pt
  }
}

\begin{tikzpicture}[my label/.style n args={2}{label={[font=\small,text=white,xshift=-0.35cm]#1:#2}}]
 
    \node[above right, inner sep=0] (image) at (0,0) {
        \includegraphics[width=\textwidth]{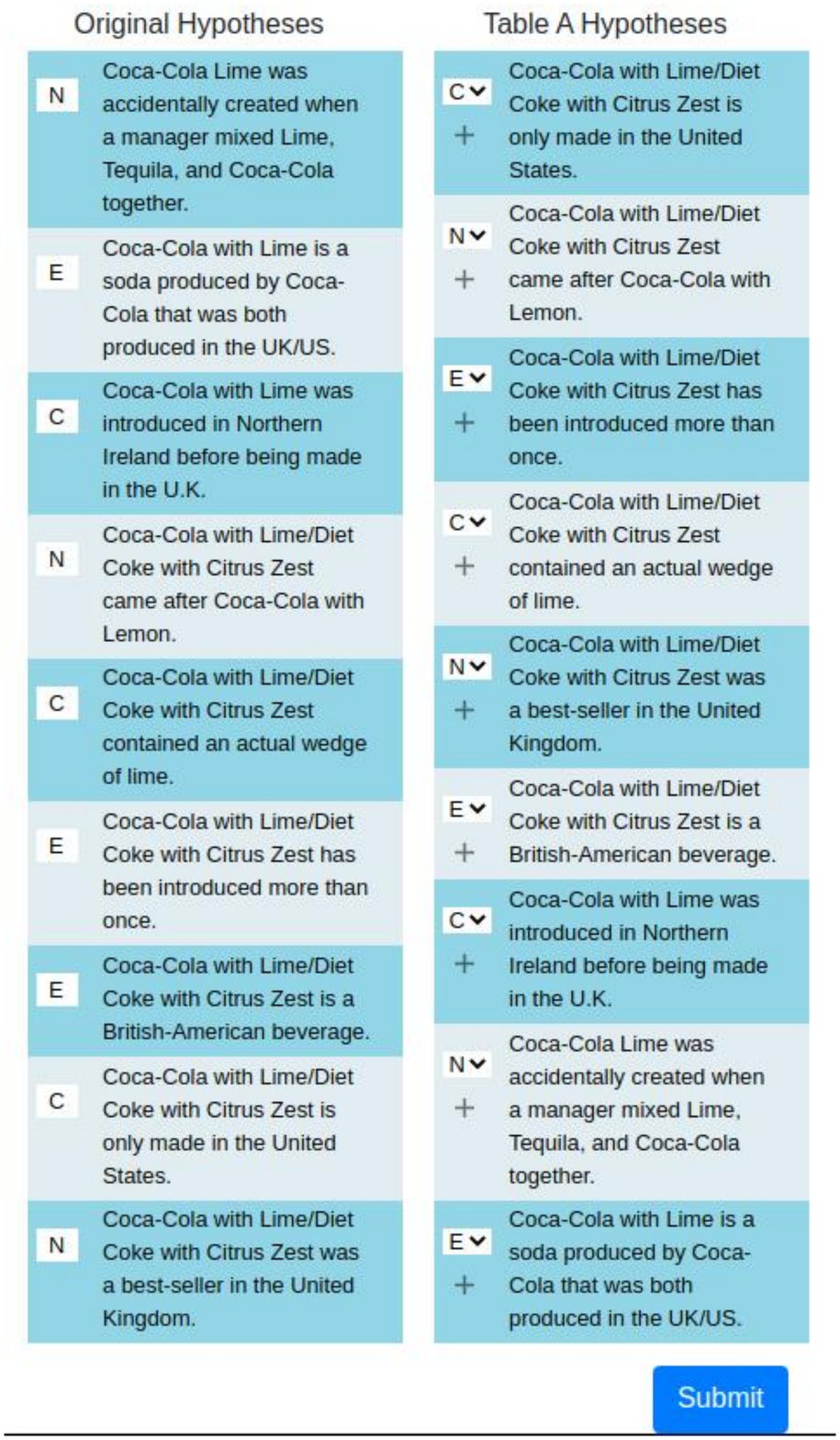}
    };
    \begin{scope}[
    x={($0.1*(image.south east)$)},
    y={($0.1*(image.north west)$)}]
     


    \draw[thick,red] (1.75,9.4) rectangle (8,9.6) node[below right,mystyle,yshift=-0.03cm,xshift=0.2cm,my label={0}{\textbf{1}}]{};
    \draw[thick,red] (1.3,7.05) rectangle (1.9,6.77) node[below left,mystyle,yshift=0.21cm,xshift=-0.55cm,my label={1}{\textbf{2}}]{};
    \draw[thick,red] (5.1,5.1) rectangle (5.8,5.4) node[below right,mystyle,yshift=0.4cm,xshift=-0.4cm,my label={0}{\textbf{3}}]{};
    \draw[thick,red] (1.9,5.3) rectangle (4.7,4.4) node[below right,mystyle,yshift=0.4cm,xshift=-2.45cm,my label={0}{\textbf{4}}]{};
    \draw[thick,red] (5.75,7.4) rectangle (8.6,6.5) node[below right,mystyle,yshift=0.5cm,xshift=0.2cm,my label={0}{\textbf{5}}]{};
    \draw[thick,red] (5.1,2.5) rectangle (5.8,2.2) node[below right,mystyle,yshift=-0.18cm,xshift=-0.36cm,my label={0}{\textbf{6}}]{};
    \node[mystyle, my label={0}{\textbf{7}}] at (6.9,0.4){};

\end{scope}
 
\end{tikzpicture}

%% file: screenshot_strategies.tex
\tikzset{
mystyle/.style={
  circle,
  inner sep=0pt,
  align=center,
  draw=black,
  fill=black,
  text=white,
  minimum size=8pt
  }
}

\begin{tikzpicture}[my label/.style n args={2}{label={[font=\small,text=white,xshift=-0.35cm]#1:#2}}]
 
    \node[above right, inner sep=0] (image) at (0,0) {
        \includegraphics[width=\textwidth]{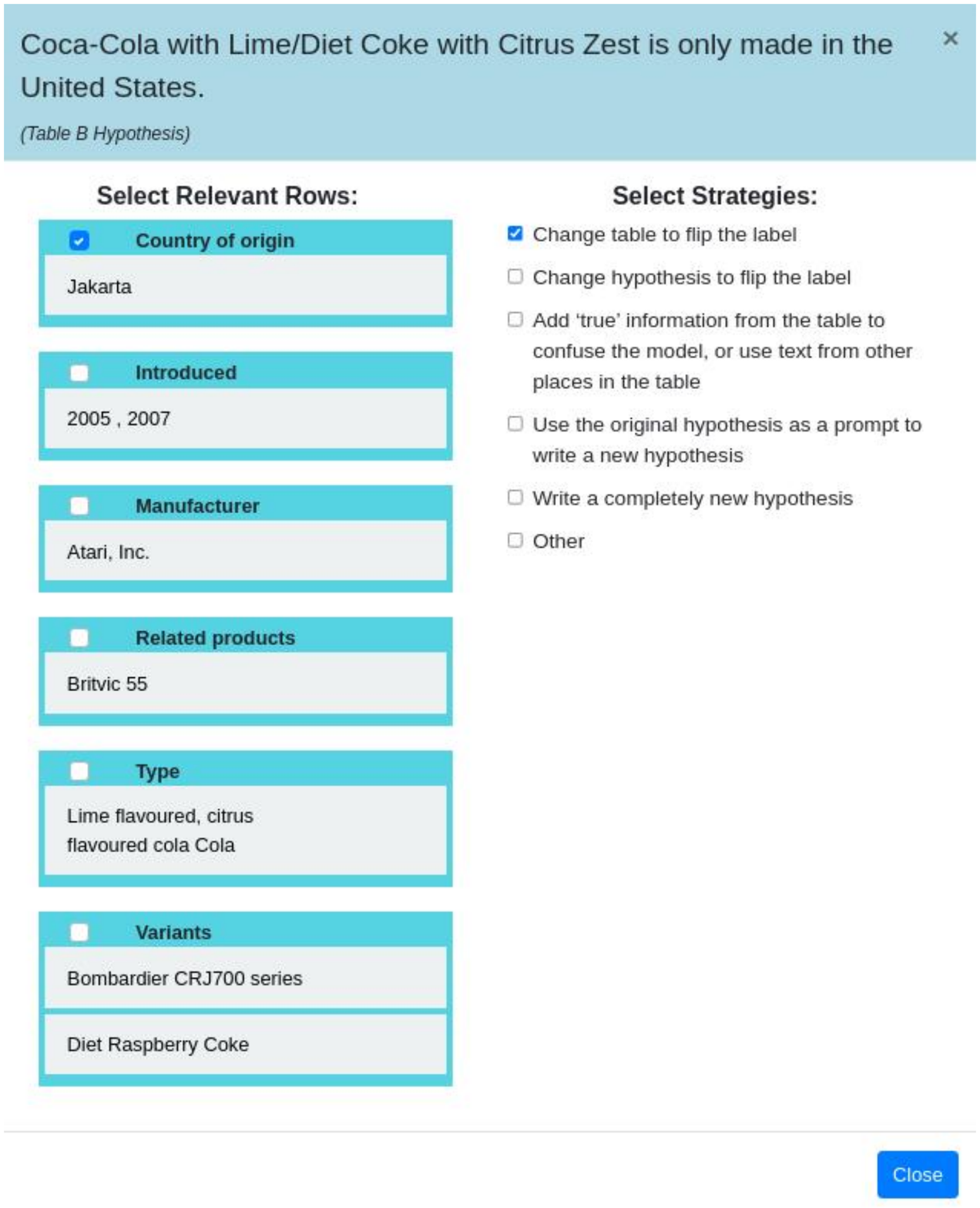}
    };
    \begin{scope}[
    x={($0.1*(image.south east)$)},
    y={($0.1*(image.north west)$)}]
     


    \node[mystyle, my label={0}{\textbf{1}}] at (2.6,8.6){};
    \node[mystyle, my label={0}{\textbf{2}}] at (2.4,8.25){};
    \node[mystyle, my label={0}{\textbf{3}}] at (0.4,7.25){};
    \node[mystyle, my label={0}{\textbf{4}}] at (0.4,5.5){};
    \node[mystyle, my label={0}{\textbf{5}}] at (8.3,7.6){};
    \node[mystyle, my label={0}{\textbf{6}}] at (7.3,6.55){};
 
\end{scope}
 
\end{tikzpicture}

%% file: casestudy_infotabs.tex
\section{Case Study on \datasetName}
\label{sec:case_study}

We use \platformName to create counterfactual data for the \datasetName dataset \cite{gupta-etal-2020-infotabs}. \datasetName is a semi-structured natural language inference dataset which consists of entity tables and human-written hypotheses. For this we sampled 47 tables with 423 table-hypothesis pairs taken from the \alphaOne set of \datasetName. \datasetName stores each table as a JSON file with key-value as attribute (column 1) - value (column 2) of the table.  For initialization we used both the \trainSet and other \alphaOne set. Including both set creates more diversity and variation in automatic initialization\footnote{As discussed earlier, annotators manually fix the constraint violations during annotation}.

\textbf{Annotation Guidelines:} Following a similar line as earlier works by \citet{ribeiro-etal-2020-beyond, Sakaguchi2020WINOGRANDEAA} for creating challenging adversarial test-sets, we guided the annotators in annotating three counterfactual table (A, B, C) for each original \alphaOne table. This ensures enough diversity and coverage in the collected counterfactual data. We advised annotators to follow the below strategies for each counterfactual table:  \begin{inparaenum}[(a)] \item For Table A: change the table in a way that it flips entail and contradict labels while keeping the hypothesis same,  \item For Table B: change the hypothesis in a way that it flips entail and contradict labels; modify the table as needed for the same, and \item For Table C: write a new but related hypotheses with similar reasoning; one can modify table as needed. \end{inparaenum}

Furthermore, for making the neutrals harder we advise annotators to modify them by adding extra \emph{`true'} information from the same table in the hypotheses. The above discussed procedure ensures that \begin{inparaenum}[(a)] \item the final label is balanced, \item hypothesis bias is destroyed by flipped label \cite{gupta-etal-2020-infotabs,chen2019tabfact}, and \item `neutrals' are now closer to `entails' in terms of lexical overlap \cite{glockner_acl18}. \end{inparaenum}. Finally, after annotation we have 109 counterfactual tables with a total of 982 table-hypothesis pairs, where Table A (141) type having 423, Table B (135) type having 405, and Table C type having 154 (52) pairs. 

\paragraph{Experiment and Analysis} To check if the annotated counterfactual data is challenging for existing model, we use \datasetName RoBERTa$_{Large}$ `para' representation and hypothesis-only model for making predictions on the original and counterfactual annotated data. Table \ref{tab:main_resutls} shows the performance result. The data was represented in `para' form, one with all rows' sentences (all Rows) and other with just relevant row sentences (coming from the annotation meta-data). 

\textbf{Performance Analysis: }Clearly the same model struggles with the counterfactual annotated data. Furthermore, better performance with relevant rows for counterfactual data shows that the model is probably using irrelevant rows tokens as artifacts for making predictions \cite{neeraja-etal-2021-incorporating}. The hypothesis-only model performance on counterfactual data is close to majority-label baselines. Additionally, human find the original and counterfactual data equally challenging, obtaining performance of $\approx$ 85$\%$ on both sets. \footnote{No difference in performance between A, B, and C type counterfactual table-example pairs, see appendix Figure \ref{fig:table_perturb_stats}}. 

\begin{table}[h]
    \small
    \centering
    \begin{tabular}{c|c|c} 
    \toprule
    \bf Model Type & \bf Original & \bf Counterfact \\ \midrule
    Majority & 33.33 & 33.33 \\
    Hypo Only & 64.32 &	44.85 \\
    All Rows & 78.91 &	61.26 \\
    Relevant Rows &  74.11 &	65.85 \\ \hdashline
    Human & 84.8 & 85.8 \\
    \bottomrule
    \end{tabular}
    \caption{\small \datasetName RoBERTa$_{Large}$ model with original and counterfactual annotated data.}
    \label{tab:main_resutls}
\end{table}

\begin{figure}[!htbp]
    \centering
    \includegraphics[width=0.93\textwidth]{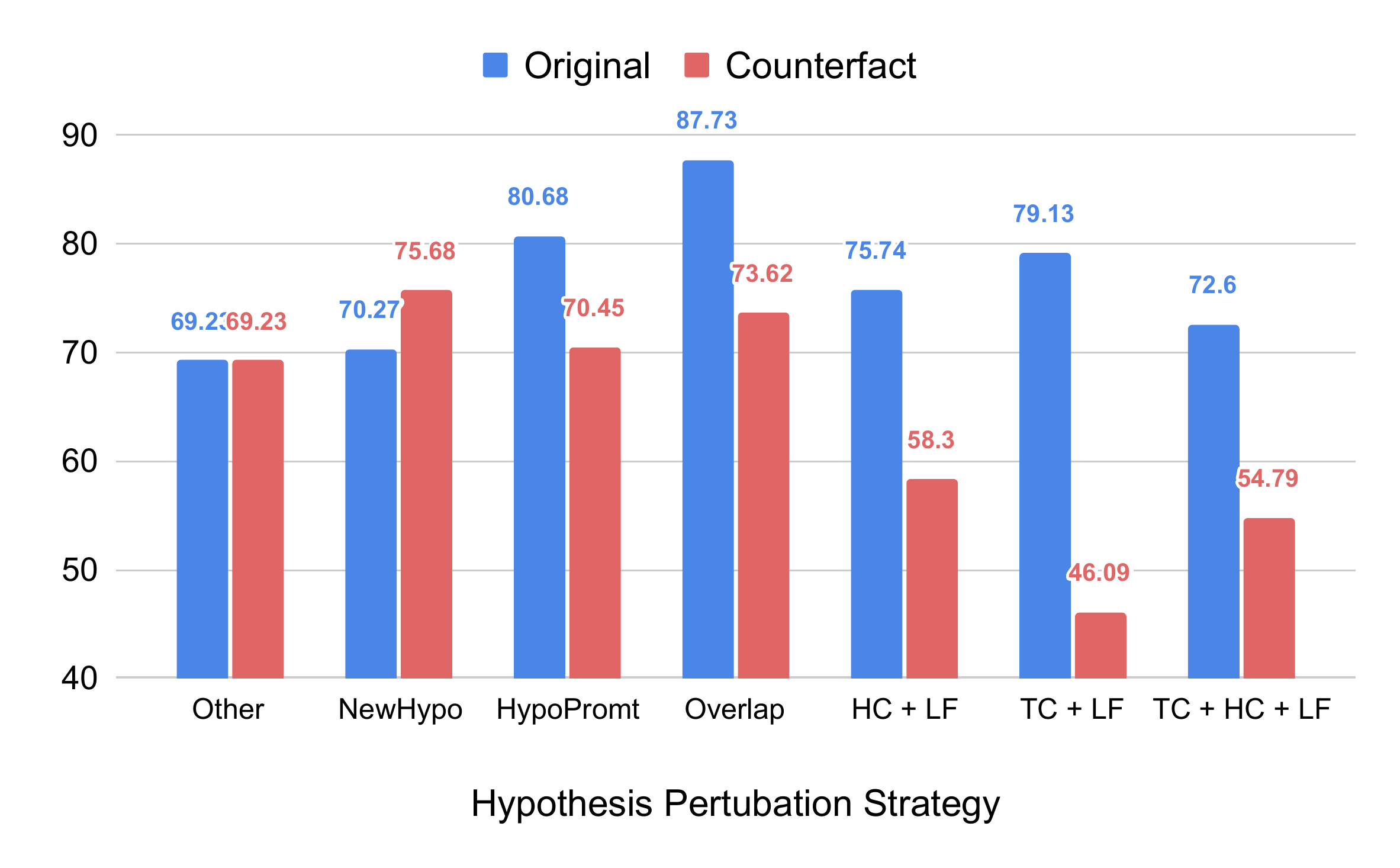}
    \includegraphics[width=0.93\textwidth]{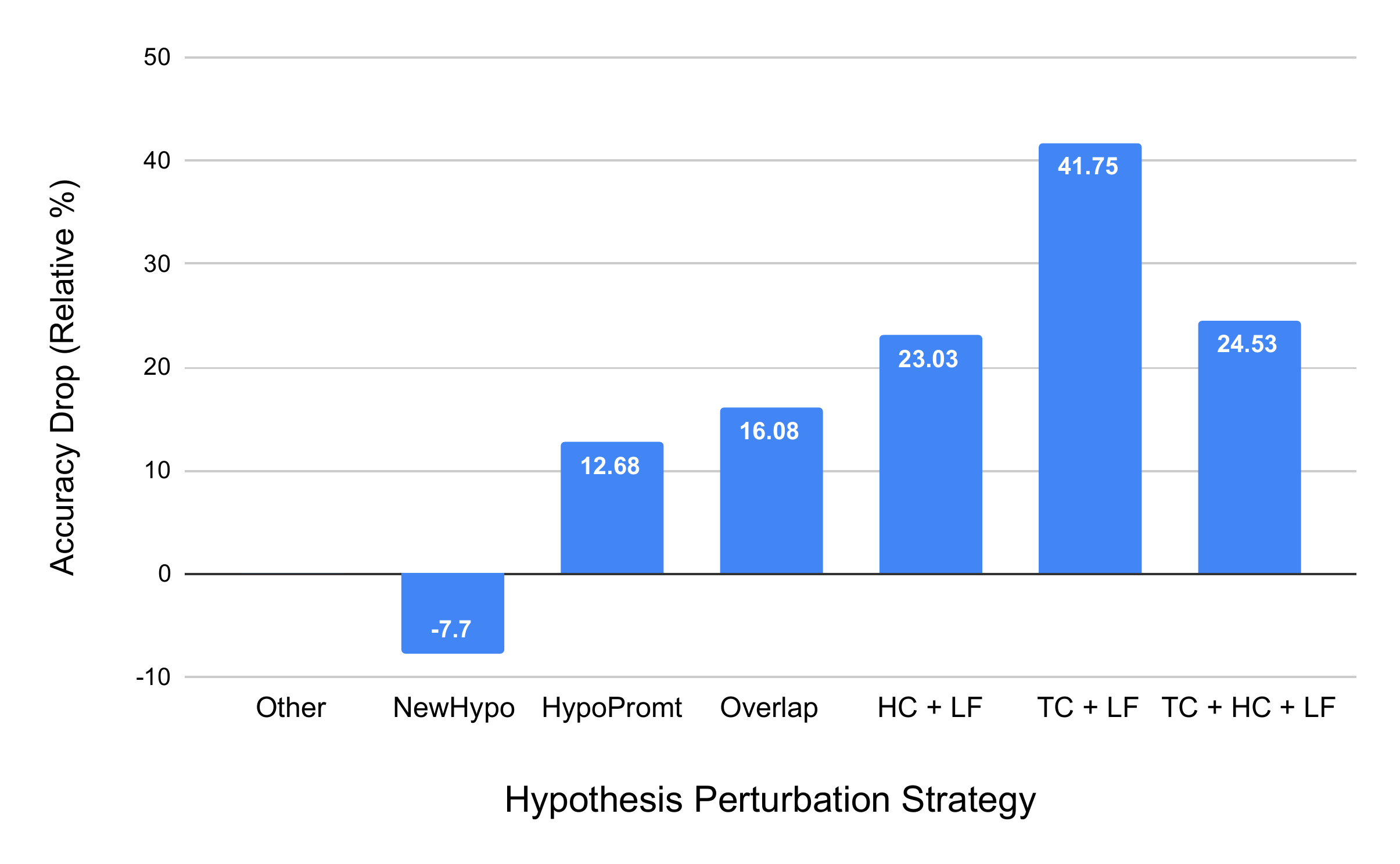}
    \caption{Performance drop after counterfactual perturbation with various strategies.}
    \label{fig:hypo_perturb_stra}
\end{figure}

\textbf{Perturbation Analysis: }We also do an analysis using the hypothesis annotation metadata to check which strategies of hypothesis modification is more effective. From Figure \ref{fig:hypo_perturb_stra}, it is evident that manual table change (TC) for Label Flip (LF) is more effective than manual hypothesis change (HC) for label flip (TF). Furthermore, strategies involving label flip is more effective then Hypothesis Prompt (Hypo Prompt) and text overlap. We suspect this is because of the ineffectiveness of hypothesis bias with flipped labels. Surprising, on new hypotheses, there is marginal performance improvement, indicating simple new data creation is an ineffective approach. Furthermore, there is no significant performance drop with any other perturbation approaches.

We also did a similar analysis on the table perturbation metadata; refer to appendix section \ref{sec:appendix_table_pert_stats} for details. We also show some qualitative example of counterfactual perturbation for each strategy in appendix section \ref{sec:appendix_examples}.

%% file: utility.tex
\section{\platformName Utility}
\label{sec:utility}

\paragraph{Main Platform: } \platformName is a tool designed particularly for annotating counterfactual tabular reasoning data, and as such, it has numerous optimizations, tools, and features that aid in the creation and collection of huge amounts of data, as well as annotating it faster and better. It allows for a broader range of tasks than, perhaps, utilizing spreadsheets or MTurk to modify such data. The drag-and-drop functionality simplifies annotation, making it simple to visualize a complicated job. In tabular form, all of the data may be examined at once. The background type validation reduces mistakes while dragging and dropping.

\paragraph{Other Task Usage:} The initialization source code, as well as the platform, are designed to be modular, so that new components can be readily added, deleted, and existing ones may be updated. For example, the ability to reorganize table parts; copying values across table triplets (in addition to cut-paste); and auto-saving work with an undo option\footnote{Currently, the Save button must be pressed manually, however it can be pressed several times at different timestamp to save even semi-completed work.}, as well as checkpoints added to reverse mistakes, may all be readily implemented.

\paragraph{Meta-Data:} The metadata collected might be used to generate challenging counterfactual adverserial test sets. For example, if a hypothesis comprises of shuffled rows from the train set and the original inference label is inverted, it might be a good candidate for evaluating NLI model overfitting. Furthermore, it is an excellent test for hypothesis artifacts if the hypothesis remains the same but the label is flipped. The hypotheses-specific rows can be used to reduce the table and explain the inference label reasoning. The marked labels can be utilized as the gold standard for existing label verification. Counterfactual tables may also be used to assess pre-trained knowledge overfitting. These are only a few examples of the numerous conceivable application situations.

We also compare and contrast \platformName with Spreadsheet on effectiveness, visual benefits, and meta-data collection aspects in appendix section \ref{sec:appendix_comparision}

%% file: future_work.tex
\section{\platformName Limitations and Future}\label{sec:future}

During our pilot study, the platform was run locally by the annotators. This was not problematic since the number of annotators was small and the tables were divided between them. We need to host our platform on a central server if we want multiple annotators to be able to make simultaneous edits for large-scale deployment. This is something which we plan to do in the near future. Lastly, the counterfactual data created by making changes had to be manually saved by pressing a button. This was done so that if the user made some mistake the original data would not be lost and the user could store the data after being satisfied with the changes. We wish to add a auto-save feature along with undo options to cater to both these scenarios.

%% file: conclusions.tex
\section{Conclusions}
\label{sec:conclusions}

We proposed \platformName, a powerful platform for visualizing semi-structured tabular data and generating counterfactual tabular perturbations. This platform allows researchers to alter tables and hypothesis phrases as well as collect accompanying metadata in order to statistically and methodically examine the shortcomings of their systems. We hope that \platformName will be useful for academics working on semi-structured data, such as tables. \platformName may also be used in non-academic industrial settings that demand table change, such as e-commerce product specification tables, financial and tax statements tables, and so on.

%% file: table_perturb_stats.tex
\section{Performance vs Perturbation}
\label{sec:appendix_table_pert_stats}

\begin{figure}[!htbp]
    \centering
    \includegraphics[width=0.96\textwidth]{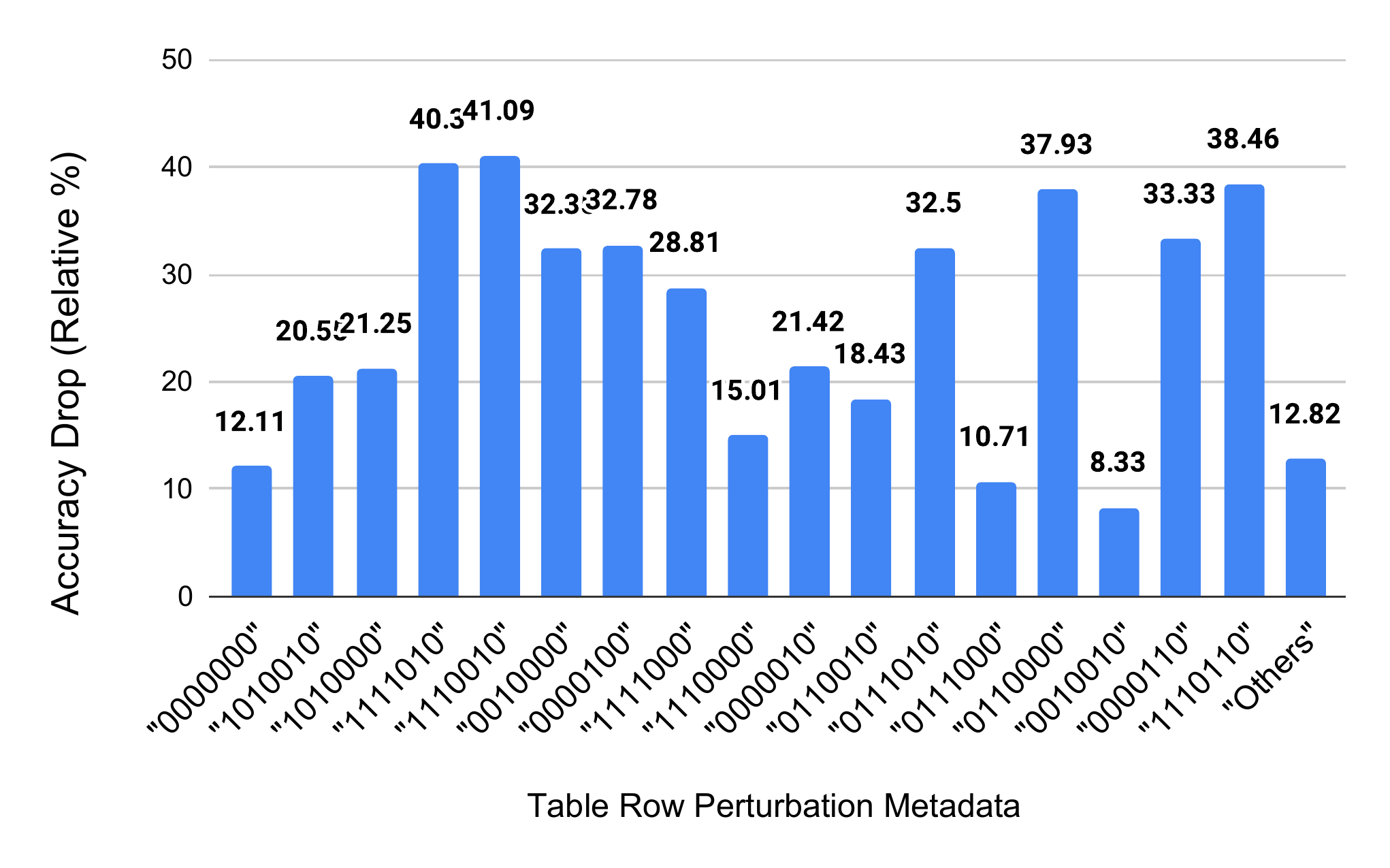}
    \includegraphics[width=0.96\textwidth]{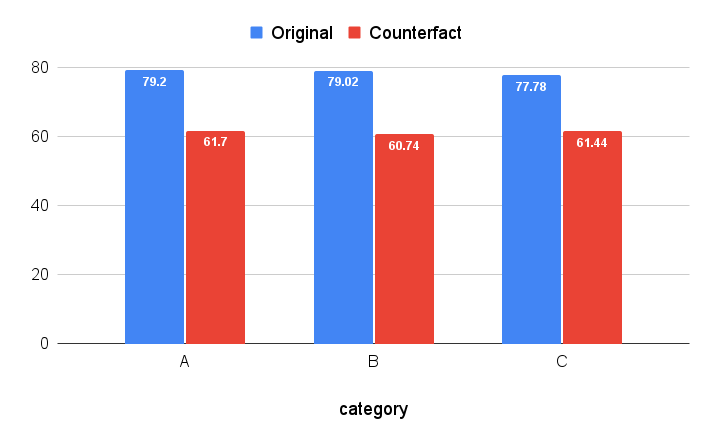}
    \caption{\small Performance drop with various counterfactual perturbation with various strategies (use the bit tables \ref{tab:fourbits} for the analysis). Performance on A,B and C type of counterfactual.}
    \label{fig:table_perturb_stats}
\end{figure}

\begin{figure}[!htbp]
    \centering
    \includegraphics[width=0.95\textwidth]{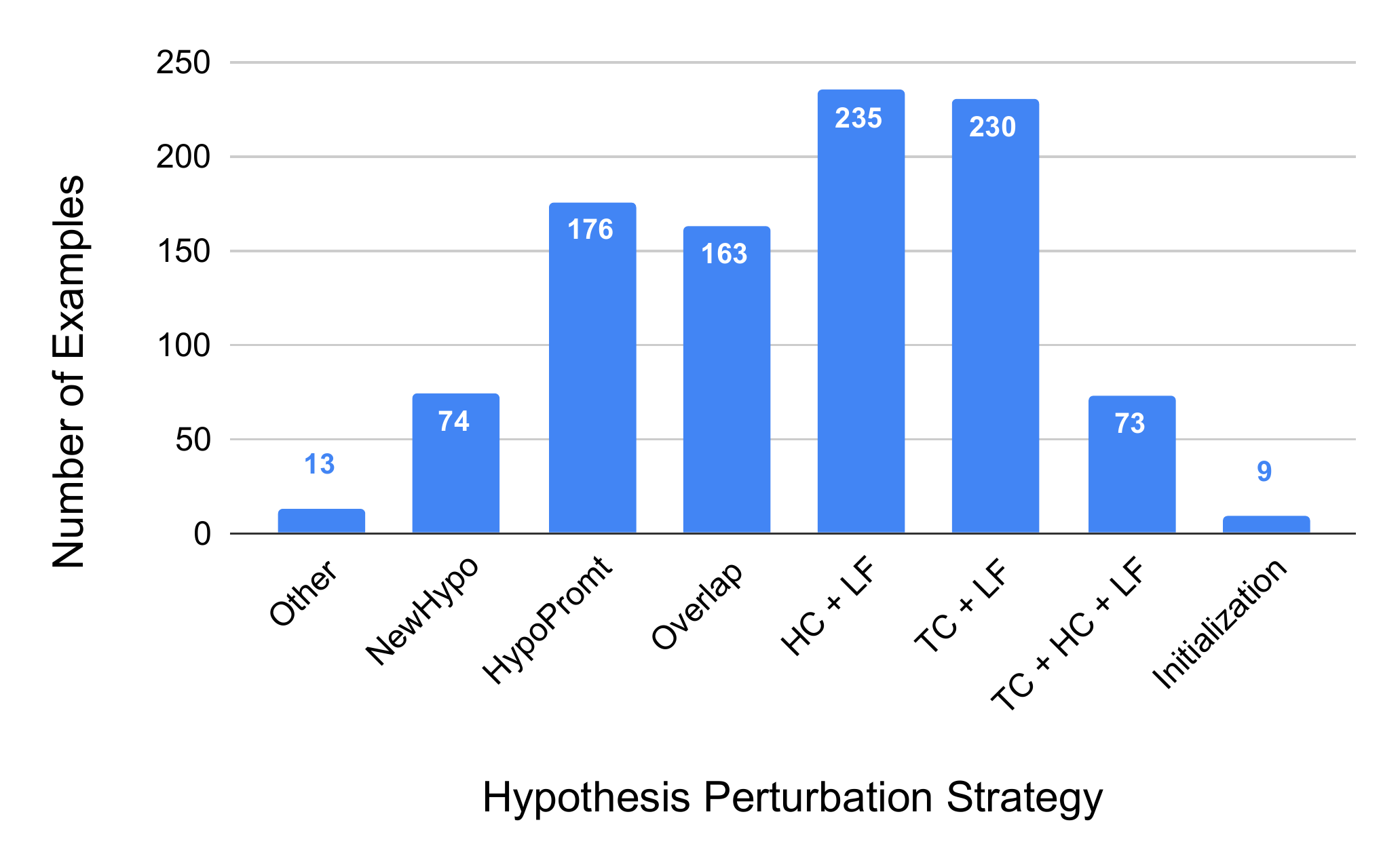}
    \includegraphics[width=0.95\textwidth]{hypo_pert_stats2.pdf}
    \caption{\small \#Example of each hypothesis perturbation strategy and relative accuracy drop for each perturbation strategy.}
    \label{fig:num_examples_hypo}
\end{figure}

Figure \ref{fig:table_perturb_stats} shows the model performance relative accuracy drop with variable table perturbation strategy. Analysing average accuracy across the A, B, and C counterfactual type, shows no significant difference. We also shows the number of examples for each hypothesis perturbation strategy and the relative accuracy drop in Figure \ref{fig:num_examples_hypo} for each strategy.

%% file: comparision.tex
\section{\platformName vs Spreadsheet}
\label{sec:appendix_comparision}

\paragraph{Effectiveness: } When utilizing spreadsheets for annotation, it becomes quite difficult and time consuming to cut(copy)-paste in numerous rows cells several times. However, the efficient drag(click)-drop with auto checking restrictions in \platformName makes it a lot easier and faster procedure. Editing and text alteration are also easier than in a spreadsheet. Our study found that it takes around 7 minutes on average to annotate a new table with 9 statements using \platformName, but the same work in the spreadsheet would take more than 40 minutes.

\paragraph{Visualisation Benefit: } Our platform's table visualisation provides a full data view on a single screen. Seeing the entire picture (table and sentences) is incredibly helpful for quality checking of annotations. It also allows the annotator to quickly follow label and hypothesis changes, which is not feasible with the spreadsheet's cell type view. 

Furthermore, having a single screen focus view on a single counterfactual table makes altering hypotheses even easy. Using this focus feature, it is straightforward to update the labels or add new information to the hypothesis. This focus view is not viable with a spreadsheet; to make appropriate alterations, one must search and navigate to each spreadsheet cell. 

Finally, the lack of scrolling required while dragging and dropping on our platform saves annotators time. To discover the relevant cells in a spreadsheet, one must execute numerous scrolling operations to the up, down, left, or right. 

Furthermore, in \platformName, the cell size is automatically set according on the underlying information, but in spreadsheet, this must be handled manually.

\paragraph{Meta-Data Collection: } \platformName also makes it simple to gather information such as methods used to change a hypothesis and rows utilized to answer each hypothesis by simply utilizing a checkbox. In a spreadsheet, this would require 9 columns of check-boxes for each table or manually writing the meta0data, which is now automatically done with a single click, thus making the process simpler, efficient and speedier.

%% file: examples.tex
\onecolumn
\section{Qualitative Counterfactual Perturbation Examples}
\label{sec:appendix_examples}

Tables \ref{table:example1}, \ref{table:example2}, \ref{table:example3}, \ref{table:example4}, \ref{table:example5} illustrate the five strategies used for counterfactual table-hypothesis perturbation. In the \textit{After} rows, the 7-bit meta-data associated with each value is also listed. The \textit{Automatic Initialisation} row explains the meaning of the first four bits of this meta-data, and the \textit{Manual Editing} row explains the last three bits.

\begin{table*}[!htbp]
\small 
\caption{\small Example using Strategy 1 \\ \textbf{Strategy}: Change table to flip label}
\centering
\begin{tabular}{p{1.8cm}|p{5cm} p{3cm} P{0.7cm} P{1cm}}
\toprule
 { }& Premise & Hypothesis & Label&Predicted\\
\hline
Before (T14) &\setlist{nolistsep}
Box Office
\begin{enumerate}  
\vspace{-0.1cm}
\item \$61.3 million
\end{enumerate}
Budget
\begin{enumerate}
\item \$26 million
\end{enumerate}
&Flatliners made over double what it cost to make at the box office. &E &E\\
\hline After (T14A) &\setlist{nolistsep}
Box Office
\begin{enumerate}
\item \$ 140.7 million (1010 010)
\end{enumerate}
Budget
\begin{enumerate}
\item \$85 million (0111 010)
\end{enumerate}
&Flatliners made over double what it cost to make at the box office.&C &E\\
\hline
Automatic\\ Initialisation & \multicolumn{4}{l}{1010: different dataset, same category, different table, same key} \\
{ }&\multicolumn{4}{l}{0111: same dataset, different category, different table, different key}\\
Manual\\ Editing & \multicolumn{4}{l}{010: value text edited}\\
\bottomrule
\end{tabular}
\label{table:example1}
\end{table*}

\begin{table*}[!htbp]
\small
\caption{\small Example using Strategy 2 \\ Strategy: Change hypothesis to flip label}
\centering
\begin{tabular}{p{1.8cm}|p{5cm} p{3cm} P{0.7cm} P{1cm}}
\toprule
 { }& Premise & Hypothesis & Label&Predicted\\
\hline
Before (T14) &\setlist{nolistsep}
Box Office
\begin{enumerate}
\item \$61.3 million
\end{enumerate}
Budget
\begin{enumerate}
\item \$26 million
\end{enumerate}
&Flatliners made over double what it cost to make at the box office &E&E \\
\hline After (T14B) &\setlist{nolistsep}
Box Office
\begin{enumerate}
\item \$ 13.3 million (1010 010)
\end{enumerate}
Budget
\begin{enumerate}
\item \$5.9 million (0110 010)
\end{enumerate}
&Flatliners made over \textcolor{red}{triple} what it cost to make at the box office.&C&E\\
\hline
Automatic\\ Initialisation& \multicolumn{4}{l}{0110: same dataset, different category, different table, same key }\\
{ }& \multicolumn{4}{l}{1010: different dataset, same category, different table, same key} \\
Manual\\ Editing & \multicolumn{4}{l}{010: value text edited}\\
\bottomrule
\end{tabular}
\label{table:example2}
\end{table*}

\begin{table*}[!htbp]
\small 
\centering
\caption{\small Example using Strategy 3 \\ \textbf{Strategy} : Add ‘true’ information from the table to confuse the model, or use text from other places in the table.}
\begin{tabular}{p{1.8cm}|p{5cm} p{3cm} P{0.7cm} P{1cm}}
\toprule
 { }& Premise & Hypothesis & Label&Predicted\\
\hline
Before (T14) &\setlist{nolistsep}
Produced by
\begin{enumerate}
\item Michael Douglas
\item Rick Bieber
\end{enumerate}
Directed by
\begin{enumerate}
\item Joel Schumacher
\end{enumerate}
&Rick Bieber put more money into Flatliners than Michael Douglas did.&N&N \\
\hline After (T14A) &\setlist{nolistsep}
Produced by
\begin{enumerate}
\item Rick Bieber (0000 100)
\item Michael Douglas (0000 100)
\end{enumerate}
Directed by
\begin{enumerate}
\item Empress Teimei (1111 000)
\end{enumerate}
 &Rick Bieber put more money into Flatliners \textcolor{red}{directed by Empress Teimei,} than Michael Douglas did&N&N\\
\hline
Automatic\\ Initialisation & \multicolumn{4}{l}{0000: same dataset, same category, same table, same key} \\
{ } & \multicolumn{4}{l}{1111: different dataset, different category, different table, different key }\\
Manual\\ Editing & \multicolumn{4}{l}{000: no change}\\
{ } &\multicolumn{4}{l}{100: copied from the original table}\\
\bottomrule
\end{tabular}
\label{table:example3}
\end{table*}

\begin{table*}[!htbp]
\small 
\caption{\small Example using Strategy 4 \\ \textbf{Strategy}: Use the original hypothesis as a prompt to write a new hypothesis}
\centering
\begin{tabular}{p{1.8cm}|p{5cm} p{3cm} P{0.7cm} P{1cm}}
\toprule
 { }& Premise & Hypothesis & Label&Predicted\\
\hline
Before (T14) &\setlist{nolistsep}
Edited by
\begin{enumerate}
\item Robert Brown
\end{enumerate}
Written by
\begin{enumerate}
\item Peter Filardi
\end{enumerate}
&Flatliners was Peter Filardi's first writing credit.&N&N\\
\hline After (T14B) &\setlist{nolistsep}
Edited by
\begin{enumerate}
\item James Newton Howard (0000 100)
\item Robert Brown (0000 000)
\end{enumerate}
Written by
\begin{enumerate}
\item Lee Beom-seon (1110 000)
\end{enumerate}
&Flatliners was \textcolor{red}{mostly edited by Robert Brown.}& N&E.\\
\hline
Automatic\\ Initialisation & \multicolumn{4}{l}{0000: same dataset, same category, same table, same key} \\
{ } & \multicolumn{4}{l}{1110: different dataset, different category, different table, same key }\\
Manual\\ Editing& \multicolumn{4}{l}{000: no change}\\
{ } &\multicolumn{4}{l}{100: copied from the original table}\\
\bottomrule
\end{tabular}
\label{table:example4}
\end{table*}

\begin{table*}[!htbp]
\small 
\caption{\small Example using Strategy 5\\ \textbf{Strategy}: Write a completely new hypothesis}
\centering
\begin{tabular}{p{1.8cm}|p{5cm} p{3cm} P{0.7cm} P{1cm}}
\toprule
 { }& Premise & Hypothesis & Label&Predicted\\
\hline
Before (T14) &\setlist{nolistsep}
Box Office
\begin{enumerate}
\item \$61.3 million
\end{enumerate}
Budget
\begin{enumerate}
\item \$26 million
\end{enumerate}
&Flatliners made 50 million over it's budget at the box office. &C&E \\
\hline After (T14C)&\setlist{nolistsep}
Box Office
\begin{enumerate}
\item US\$85.4 million (December 2017) (0111 000)
\end{enumerate}
Budget
\begin{enumerate}
\item \$26 million (0000 000)
\end{enumerate}
&\textcolor{red}{Flatliners costed around \$25 million in making and was a hit.} & E & C \\
\hline
Automatic\\ Initialisation &\multicolumn{4}{l}{0000: same dataset, same category, same table, same key}\\
{ } & \multicolumn{4}{l}{0111: same dataset, different category, different table, different key}\\
Manual\\ Editing & \multicolumn{4}{l}{000 : no change}\\
\bottomrule
\end{tabular}
\label{table:example5}
\end{table*}